# A computer vision method to estimate ventilation rate of Atlantic salmon in sea fish farms

Lukas Folkman[a,b,c,*], Quynh LK Vo[a,c], Colin Johnston[d], Bela Stantic[b], Kylie A Pitt[a]

[a] Coastal and Marine Research Centre, Australian Rivers Institute, Griffith University, QLD 4222, Australia

[b] Institute for Integrated and Intelligent Systems, School of Information and Communication Technology, Griffith University, QLD 4222, Australia

[c] Blue Economy Cooperative Research Centre, Launceston, TAS 7248, Australia

[d] Tassal Group, Hobart, TAS 7000, Australia

* Corresponding author

## Abstract

The increasing demand for aquaculture production necessitates the development of innovative, intelligent tools to effectively monitor and manage fish health and welfare. While non-invasive video monitoring has become a common practice in finfish aquaculture, existing intelligent monitoring methods predominantly focus on assessing body condition or fish swimming patterns and are often developed and evaluated in controlled tank environments, without demonstrating their applicability to real-world aquaculture settings in open sea farms. This underscores the necessity for methods that can monitor physiological traits directly within the production environment of sea fish farms. To this end, we have developed a computer vision method for monitoring ventilation rates of Atlantic salmon (*Salmo salar*), which was specifically designed for videos recorded in the production environment of commercial sea fish farms using the existing infrastructure. Our approach uses a fish head detection model, which classifies the mouth state as either open or closed using a convolutional neural network. This is followed with multiple object tracking to create temporal sequences of fish swimming across the field of view of the underwater video camera to estimate ventilation rates. The method demonstrated high efficiency, achieving a Pearson correlation coefficient of 0.82 between ground truth and predicted ventilation rates in a test set of 100 fish collected independently of the training data. By accurately identifying pens where fish exhibit signs of respiratory distress, our method offers broad applicability and the potential to transform fish health and welfare monitoring in finfish aquaculture.

# Highlights

- A method based on object detection and tracking algorithms was developed to quantify fish ventilation rates in aquaculture.
- Training and evaluation data was collected from the production environment of commercial sea fish farms.
- Specific data curation strategies contributed to training a highly accurate fish head detection model.
- The method has the sensitivity to identify fish pens with respiratory distress, as demonstrated in videos recorded during commercial aquaculture operations.

# 1 Introduction

Monitoring the health and welfare of farmed fish is one of the core tasks of the aquaculture industry. Maintaining fish in optimal condition is imperative for the industry to avoid stock losses, maximise overall production, and minimise production costs. Computer vision has emerged as a powerful, non-invasive tool for monitoring in aquaculture (Barreto et al., 2022). However, most computer vision methods developed for the aquaculture industry have only been evaluated in fish tanks, where the environmental conditions such water turbidity and available light can be controlled. While some applications such as monitoring feeding efficiency or assessing body condition have already been deployed to commercial farms (Rishi et al., 2019), there is a lack of methods that can monitor physiological traits, such as ventilation rates (Høgstedt et al., 2025; Zheng et al., 2014), directly within the production environment of commercial sea fish farms.

The health of farmed fish is usually assessed through manual inspections of a sub-sample of fish within a pen (Mitchell et al., 2012; Parsons et al., 2001). This procedure requires fish to be removed from pens, anaesthetised, visually inspected, and returned to pens. Manual health inspections are, therefore, stressful for fish and are labour-intensive and expensive for industry. Depending on how frequently fish are inspected, manual health inspections may not detect problems within a timeframe before lag indicators, such as feed intake, become apparent. Continuous and near real-time monitoring of fish is needed, therefore, to enable farmers to identify and respond to problems on the farm before they escalate.

Rapid advances in technologies are changing the way farmers monitor fish and are enabling problems to be detected sooner (Barreto et al., 2022). For example, MinION nanopore sequencing enables salmonid viruses to be sequenced in the field within hours instead of the weeks taken for standard laboratory-based PCR assays (Gallagher et al., 2018). Biosensors can be surgically attached to sentinel fish within a pen and either store or, ideally, transmit real-time information on behaviour (e.g. swimming velocity and operculum movements) and metabolism (Calduch-Giner et al., 2022). Both these technologies, however, are expensive, and the surgical interventions associated with implanting biosensors induce stress on the fish (Calduch-Giner et al., 2022; Gallagher et al., 2018).

Video cameras deployed within pens, that stream live videos to staff on shore, enable real-time monitoring and are routinely used in production of high-value species, such as salmonids. The video streams can be analysed using computer vision methods, which addresses the shortcomings of both manual assessments and devices that require surgical interventions: computer vision offers continuous, real-time, non-invasive, and cost-effective option for monitoring fish health and welfare.

Both classical computer vision and deep-learning algorithms have been developed for smart monitoring in aquacultural facilities (H. Liu et al., 2023; S. Zhao et al., 2021; Zion, 2012). These methods have been applied to assess fish behaviour in relation to feeding (Z. Liu et al., 2014; Måløy et al., 2019; Wei et al., 2021; L. Zhang et al., 2022; J. Zhao et al., 2017; Zhou et al., 2019), swimming (Huang et al., 2022; Koh et al., 2023; Pinkiewicz et al., 2011), and abnormal activity (X. Li et al., 2022; Y. Li et al., 2024; H. Wang et al., 2022; Z. Wang et al., 2023; Yang et al., 2024); to count fish (H. Zhang et al., 2023; Z. Zhang et al., 2024; Zhu et al., 2024); to detect feed pellets (Hu et al., 2021; Parsonage & Petrell, 2003; Xu et al., 2023); or to estimate ventilation rates (Høgstedt et al., 2025; Zheng et al., 2014).

The ventilation rate of fish, also referred to as respiratory or breathing rate, is a key indicator of stress that is routinely quantified in laboratory studies and qualitatively assessed by staff monitoring video streams from fish farms. Ventilation rates typically increase in response to changes in environmental conditions, such as exposure to ammonia (Knoph, 1996), disease (Leef et al., 2005), crowding (Erikson et al., 2016) and injuries, such as those inflicted by jellyfish. Fish ventilate their gills using a combination of buccal and opercular movements (Hughes & Shelton, 1958). When fish open their mouths and abduct their operculum, the volume of the buccal and opercular cavities increases, which lowers pressure and draws water inwards. As the mouth closes and operculum adducts, pressure increases, and water is expelled through the opercular valve. An alternative to active pumping of water, called ram ventilation, occurs when the mouth remains open while swimming or residing in a fast water current (Roberts, 1975). The movements of the mouth and operculum provide strong visual cues for developing computer vision methods to quantify fish ventilation rates.

Zheng et al. (2014) introduced a method for measuring respiratory (ventilation) rhythms in Japanese rice fish (*Oryzias latipes*) for water quality monitoring, but this method cannot be adopted for aquaculture purposes as it relies on observing changes in colour around the gill region, which occur only in particular fish species. Høgstedt et al. (2025) proposed a method for estimating the breathing (ventilation) rate of Atlantic salmon. The method assigns a unique identity to each monitored fish, necessitating training of a fish identity classification model before ventilation rates can be assessed. The approach was evaluated using a small cohort of seven individuals housed in a 1,000-litre tank. In summary, the existing computer vision approaches for estimating fish ventilation rates were not designed for nor evaluated in the highly variable conditions of sea fish farms, where water turbidity, camera motion, and lighting can degrade the video quality. Moreover, the existing methods do not scale effectively to commercial sea pens, which typically house 20,000–100,000 fish, limiting their applicability in large-scale aquaculture operations.

To address these limitations, we developed and thoroughly evaluated a computer vision approach for quantifying fish ventilation rates from videos recorded in commercial sea fish farms. The approach is widely applicable without a need of specialised equipment and could be adapted for a variety of finfish species. Our method is based on a fish head detection model combined with multiple object tracking to create temporal sequences (tracks) of fish swimming across the field of vision of the underwater video camera. We extensively tested our method using real-world, production videos collected across a number of Atlantic salmon farms in South East Tasmania. The proposed method achieved an average precision of 95.8% and 93.2% for detecting fish with open and closed mouths, respectively, and proved efficient in tracking and estimating the ventilation rate with a Pearson correlation coefficient of 0.82 between the ground truth and predicted ventilation rates of 100 fish collected independently of the training data. When the method was used to assess ventilation rates of four fish populations of which two showed signs of increased respiratory effort, the median ventilation rate proved to be a robust indicator of population-scale ventilation rates. These results demonstrate the applicability of the proposed method to real-world aquaculture settings in open sea farms. By reliably identifying pens where fish show signs of respiratory distress, our method could reduce costs and animal handling associated with routine health checks by aiding the prioritisation of pens for inspections. This early detection capability not only improves operational efficiency but also has a positive impact on fish health and welfare. By enabling quantitative, accurate, and non-invasive monitoring of ventilation rates—a critical physiological trait linked to various environmental and health conditions— our method has a potential to transform fish welfare management in aquaculture.

## 2 Materials and methods

We developed a computer vision method for estimating fish ventilation rates (**Fig. 1**). The method was designed and developed specifically for the production environment of sea fish farms (**Fig. 1a**). We trained a YOLOv8 object detection model (Jocher et al., 2023; Redmon et al., 2016) to detect fish heads and classify the mouth states as "open", "closed", or "dropped jaw" (**Fig. 1b**). Next, multiple object tracking algorithm BoT-SORT (Aharon et al., 2022; Y. Zhang et al., 2022) was used to create temporal sequences (tracks) of the detected fish heads (**Fig. 1c**). Each fish track and the associated predicted mouth states served as a ventilation sequence from which the fish ventilation rate was estimated (**Fig. 1d**).

### *2.1 Calculation of ventilation rate*

The ventilation rate was defined as the number of times a fish opens and closes its mouth per minute when not feeding. To this end, each detected fish track (temporal sequence of mouth states assigned to a single fish individual) was processed as follows. Tracks in which more than half of the mouth states were classified as "dropped jaw" (an anatomical deformity that prevents the fish from closing their mouth) were removed outright. The remaining detections classified as "dropped jaw" were treated as missed detections as they were likely misclassified instances of open or closed mouths. Missed detections occurring in a single frame were imputed by randomly choosing between the previous and subsequent mouth state. Since our manually annotated datasets contained no occurrence of open or closed mouth sequences of a single frame duration, we discarded all tracks that contained […]-open-*closed*-open-[…] instances because they likely contained misclassified fish heads. A more liberal treatment was applied to tracks containing […]-closed-*open*-closed-[…], in which case we converted the singleton open mouth state to closed mouth. This correction was implemented to counteract the tendency of the model to overpredict the open mouth state due to the overrepresentation of open mouth labels in the training dataset.

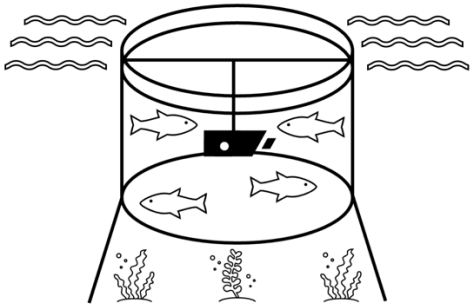
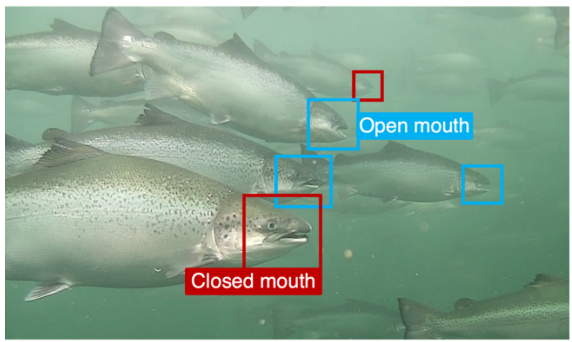
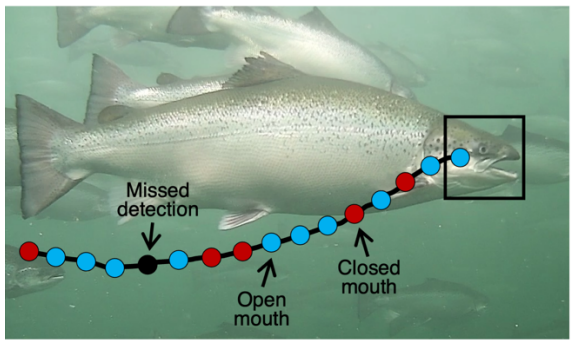
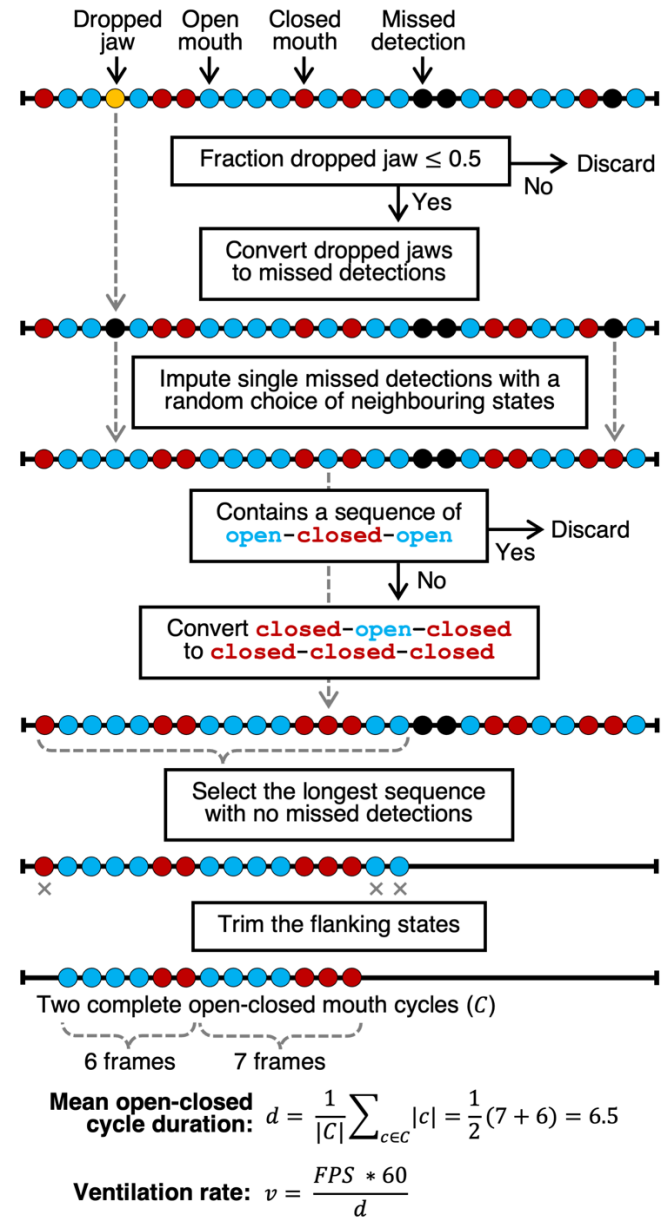

**Fig. 1.** Illustration of the proposed method. The method was designed and developed specifically for processing videos recorded with underwater cameras deployed in the production environment of sea fish farms **(a)**. Object detector (YOLOv8) was used to detect fish heads and classify them based on the mouth state as open mouth, closed mouth, or dropped jaw **(b)**. Multiple object tracker (BoT-SORT) was used to create a temporal sequence (track) of mouth states associated with the same individual **(c)**. Each fish's mouth track underwent quality control, during which missed detections and potential misclassifications of mouth states were identified and corrected, and the ventilation rate was then estimated using the longest sequence of complete open-closed mouth cycles **(d)**.

After the above data cleaning, the longest sequence with no missing detections was selected from each track. We trimmed the flanking states (as these were observed incompletely and could bias the estimates) and retained only sequences that contained at least one open-closed (or closed-open) cycle. For each track, the duration of the open-closed cycle was calculated as the mean duration (number of frames) of adjoint open-closed or closed-open sequences (depending on which state was observed first). Finally, given the video's frame rate per second ($FPS$), ventilation rate (per minute) was calculated as:

$$\text{ventilation rate [cycles per minute]} = \frac{60 \times FPS}{\text{open-closed cycle duration [number of frames]}}$$

*2.2 Datasets*

*Fish head detection datasets*

We collected 49 videos across 13 Atlantic salmon farms (35 different pens) in South East Tasmania between 19/10/2022 and 06/10/2023 during the routine operation of the fish pens containing 58,901 fish on average (ranging from 24,093–115,359) of an average weight of 2.7 kg (ranging from 1.0 to 6.2 kg). The camera was positioned at an average depth of 4.9 m (ranging from 2.6–7.7 m) pointing sideways. The collected videos had a resolution of 1280×960 pixels and a frame rate of 30 frames per second. The video durations were in the range of 1–60 minutes with a mean duration of 35 minutes and 43 seconds. For training, validation, and testing of the fish head detection model, we used 33, 12, and 4 videos, respectively (**Fig. 2a**). From each video, we manually selected frames (images) which included fish with their mouths visible, and bodies positioned mostly transversely. Computer Vision Annotation Tool (CVAT) (Sekachev et al., 2020) was used to aid manual annotation of the fish heads. Each annotation comprised a bounding box that encompassed the head of the fish (but excluded the operculum) and one of the following labels: open, closed, or "dropped jaw" (an anatomical deformity that prevents determining if the mouth was open or closed).

The annotation process included two additional considerations: (1) identification of fish which did not open their mouth sufficiently wide and (2) masking of fish heads that were deemed not suitable for annotation (**Fig. 2b**). We used the fish heads with narrow mouth openings to test if including more annotations of open mouths for training, albeit with a narrow mouth opening, was beneficial or detrimental to the accuracy of the trained model. Similarly, we tested whether masking fish heads that were deemed not suitable for annotations was beneficial or detrimental. Heads were masked by blurring a tightly fit rectangular box on the fish' head. Blurring was implemented using the Gaussian filter with a radius of 10 with the Python Imaging Library (Pillow) (Clark & others, 2023).

The training, validation, and test datasets contained 9,400, 3,005, and 1,373 annotated fish heads, respectively (**Table 1**). When fish heads with narrow mouth openings were included, the training dataset contained another 1,754 instances of fish heads with an open mouth. There was a significant imbalance among the three different categories with the open, closed, and dropped jaw labels comprising 73%, 22%, 4% instances on average, respectively. Fish heads of the same individual in different poses were frequently included. Based on the median duration it took a fish to swim across the field of view of the camera, we conservatively estimated that the total of 13,778 annotated fish heads belonged to at least 3,725 different fish individuals.

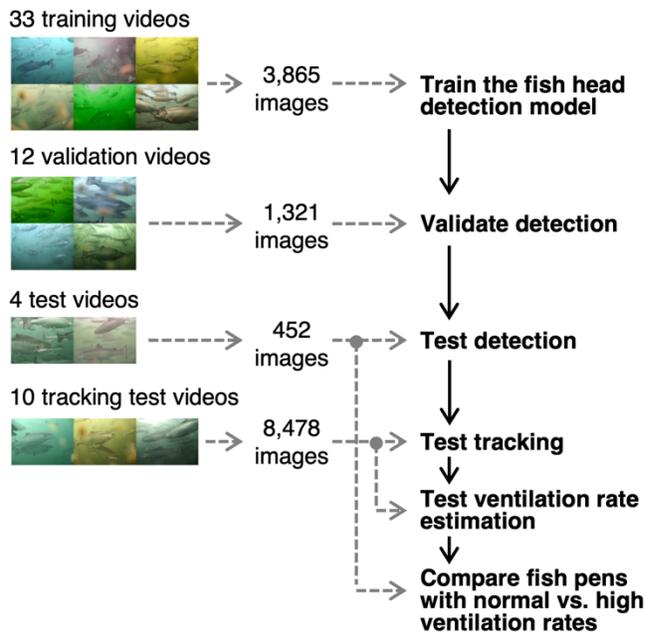
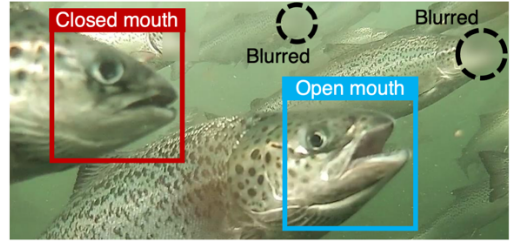
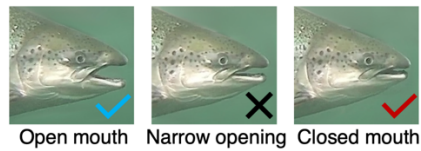

**Fig. 2.** Depiction of the data collection, annotation, and curation. A diverse set of 59 videos was collected for developing and evaluating the proposed method, with no overlap of video sources used for training, validation, and testing **(a)**. To develop a highly accurate object detector of fish mouth states, a three-stage annotation processed was used **(b)**.

*Fish tracking and ventilation rate datasets*

We collected ten videos, which were used to create a fish tracking dataset to test the accuracy of the tracking and ventilation rate estimation algorithms (**Table 1**). The videos were recorded in ten different salmon farms in South East Tasmania (three of which were not part of our training and validation datasets) between 03/10/2023 and 05/01/2024. From each video we annotated fish heads and mouth states across sequential video frames for 10 selected fish individuals that spent enough time in the field of view of the camera, while leaving all other fish in the selected frames unannotated. Thus, each track defined a complete ventilation sequence of open and closed mouth states, yielding a tracking test dataset containing 8,478 annotated fish heads of 100 different fish individuals, with fish track lengths ranging from 18 to 226 frames (median 69, mean 84.78).

Additionally, to test the methods applicability and sensitivity to an aquaculture industry relevant task, we used videos from two sea pens containing fish with normal ventilation rates and two sea pens containing fish with elevated (high) ventilation rates. The ground truth labels (normal vs high) for these videos were provided by two different aquaculture industry professionals.

**Table 1.** Datasets used for development and testing of the ventilation rate model.

| Dataset | Videos | Images | Open mouths | Closed mouths | Dropped jaws | Fish | Tracks[c] |
|---|---|---|---|---|---|---|---|
| Training | 33 | 3,865 | 6,780 | 1,995 | 625 | 2,727[b] | 0 |
| Validation | 12 | 1,321 | 2,158 | 677 | 170 | 669[b] | 0 |
| Test (detection) | 4[a] | 452 | 1,041 | 316 | 16 | 329[b] | 0 |
| Test (tracking) | 10 | 8,478 | 5,877 | 2,601 | 0 | 100 | 100[d] |

[a] Two videos were labelled as containing fish with normal ventilation rates and two videos with high levels of ventilation rates.
[b] The number of fish was approximated based on the number of annotated boxes at least 69 frames apart from each other.
[c] A track refers to a set of annotations for a specific fish individual across sequential video frames. Each track defined a complete ventilation sequence of open and closed mouth states.
[d] Track length ranged from 18 to 226 frames with a median of 69 and a mean of 84.78 frames.

## 3 Experimental settings

### 3.1 Fish head detection model

The fish head detection model was implemented with YOLOv8 from the Ultralytics package (Jocher et al., 2023), which is based on the YOLO (Redmon et al., 2016) object detection algorithm. The training of the fish head detection model was initiated using the weights of the YOLOv8 large architecture containing 43.7 million parameters pretrained on the COCO dataset (Lin et al., 2014). The model was trained for 100 epochs using a batch size of 10 images and the default maximum input size of 640×480 pixels. For all other training parameters the default values were used as we observed a good training convergence and high validation set accuracy. The default parameters included the following randomly applied data augmentations: horizontal flipping, scale and colour jittering, geometric translation, and mosaic (tiling of four training images in a 2×2 grid). A model selection procedure was used to avoid over-fitting the training dataset. The validation set $mAP_{50-95}$ was calculated at the end of every epoch and used to select the best model for implementing the ventilation rate estimation algorithm.

During inference, the input size was set to 640×480 pixels, non-maximum suppression (NMS) threshold to 0.7 (the default value), and maximum number of detections per image to 100 (higher than observed in any of the training set images).

The detection models were trained and evaluated in three replicates with random seeds of 1, 2, and 3 using NVIDIA A100 graphical processing units (GPUs). We always used the same reproducible environment with Python 3.9.18, Pytorch 1.13.0, CUDA 11.6, Ultralytics 8.0.171, NumPy 1.25.2, Pycocotools 2.0.7, OpenCV 4.8.0.76, and Pillow 9.4.0.

### 3.2 Fish tracking algorithm

We chose BoT-SORT as the tracking-by-detection algorithm because it supports camera motion compensation. This is particularly important in the sea farm scenario where the camera can exhibit significant displacement during operation due to water movement. To increase the number of detections associated with tracks, we adjusted default values of the following tracking hyper-parameters: the new track thresholds was lowered to 0.5 (from 0.6) and the detection association threshold was lowered to 0.7 (from 0.8). The default values of the tracking buffer (30) and the low and high confidence threshold (0.1 and 0.5) were used. This means that all detections with a confidence threshold less than 0.1 were excluded from tracking and ventilation rate estimation.

### 3.3 Evaluation metrics

*Fish head detection evaluation*

The fish head detection model was evaluated with the mean average precision (mAP) metric, calculated according to the COCO evaluation standard. Precision ($P$) represents the proportion of correct detections among all detections, and recall ($R$) represents the proportion of correct detections among all (ground truth) objects:

$$P = \frac{TP}{TP + FP} \qquad R = \frac{TP}{TP + FN}$$

where $TP$, $FP$, and $FN$ are the number of true positives, false positives, and false negatives, respectively. A correct detection was defined for a given intersection over union (IoU) threshold, which is the ratio of the intersection of the ground truth and predicted bounding boxes' areas to their combined areas. The average precision (AP) for a given object category (class) $k$ was interpolated across 101 recall values (ranging from 0 to 1 in 0.01 increments):

$$\text{AP}(k) = \frac{1}{101} \sum_{r \in \{0.0, \dots, 1.0\}} \max_{\tilde{r} \geq r} p(\tilde{r})$$

where $p(\tilde{r})$ is the precision value at the recall value $\tilde{r}$. Finally, mAP was the mean of AP values across the three object categories (open, closed or dropped jaw; denoted as $C$):

$$\text{mAP} = \frac{1}{|C|} \sum_{k \in C} \text{AP}(k)$$

We used mAP at the IoU threshold of 0.5 (mAP$_{50}$) as the main evaluation metric for the fish head detection model. Additionally, mAP$_{50-95}$ was used for implementing model selection, calculated as the mean of mAPs at IoU thresholds ranging from 0.50–0.95 in 0.05 increments.

*Fish head tracking evaluation*

The performance of our model to effectively track fish heads and estimate ventilation rates was evaluated using the fish tracking dataset that included 100 manually annotated fish tracks. Two aspects of the tracking performance were evaluated – association and detection accuracy. The tracking association accuracy assessed how well the tracking algorithm maintained the identity consistently over time, i.e. the association accuracy measured the correctness of object identity assignments across frames, independent of detection errors. To this end, each ground-truth track ($t^{GT}$) was matched with a single detected track ($t^{DT}$) to create one-to-one pairing such that the tracking association accuracy ($A$) was maximised:

$$A = \max_{k,l} \sum_i 1 \text{ if } IoU\left(t^{GT}_{k+i}, t^{DT}_{l+i}\right) \geq 0.33 \text{ else } 0$$

The tracking detection accuracy assessed if the open and closed mouth states were correctly classified using per-class recall and precision ($R$ and $P$, defined in Section 3.4.1) at the IoU threshold of 0.33 and confidence threshold of 0.1.

*Ventilation rate estimation evaluation*

To quantify the agreement between the ground truth ventilation rates and the predicted ventilation rates, Pearson correlation coefficient ($r$) and mean absolute error ($MAE$) were used:

$$r = \frac{\sum_{i=1}^{N}(VR_i - \text{mean}(VR))(\widetilde{VR}_i - \text{mean}(\widetilde{VR}))}{\sqrt{\sum_{i=1}^{N}(VR_i - \text{mean}(VR))^2} \sqrt{\sum_{i=1}^{N}(\widetilde{VR}_i - \text{mean}(\widetilde{VR}))^2}}$$

$$MAE = \frac{1}{N} \sum_{i=1}^{N} |\widetilde{VR}_i - VR_i|$$

where $VR_i$ and $\widetilde{VR}_i$ are the ground truth and predicted ventilation rates of the $i$-th fish individual in the fish tracking dataset, respectively, and $N$ is the number of fish with at least one observed open-closed cycle in the tracking test dataset (ventilation rate is undefined for fish that do not open or close their mouth during the time spent in the field of view of the camera).

## 4 Results

### *4.1 Detecting and classifying fish heads*

The fish head detection model trained on the baseline training dataset resulted in high validation mAP$_{50}$ of 88.5%, but the AP$_{50}$ values for the closed mouth (88.0%) and dropped jaw (81.5%) categories were considerably lower than for the prevailing category of fish with open mouth (96.1%). After removing annotations of fish heads with narrow mouth openings and blurring the fish heads in the background, the closed mouth and dropped jaw AP$_{50}$ increased to 95.7% and 87.0%, improving by 7.7 and 5.5 percentage points, respectively, compared to training with the baseline dataset (**Fig. 3**). The overall mAP$_{50}$ of the fish head detection model increased to 93.1%, an improvement of 4.6 percentage points.

To ensure that the training procedure converged to an optimal solution, which can generalise well to images outside of the training dataset, we monitored the validation loss and validation mAP$_{50–95}$ metrics (**Fig. 4**). Both metrics showed consistent trends across all three model replicates and plateaued within the first 50 training epochs, confirming that a total of 100 training epochs was sufficient. The best-performing model for each replicate was selected based on the highest validation mAP$_{50–95}$. As a result, models trained for 92, 51, and 84 epochs were chosen for replicates 1, 2, and 3, respectively.

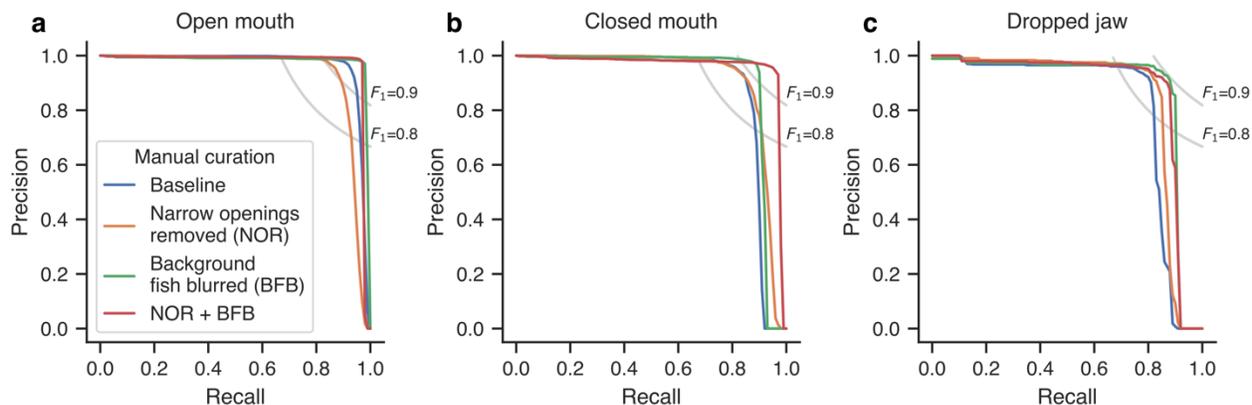

**Fig. 3.** Precision-recall curves for detecting open mouth (**a**), closed mouth (**b**), and dropped jaw (**c**), comparing different annotation strategies using the validation dataset. The grey lines depict $F_1$-score thresholds; $F_1$ is the harmonic mean of precision and recall.

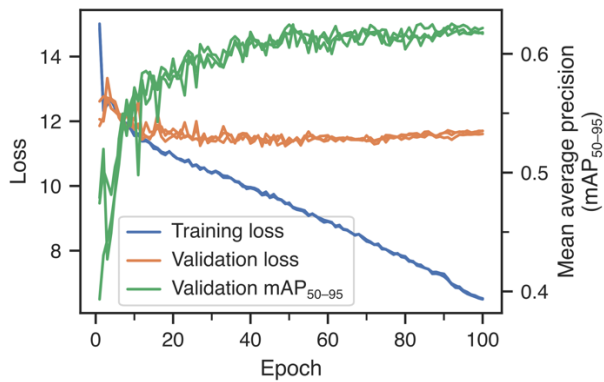

**Fig. 4.** Training loss, validation loss, and validation $mAP_{50–95}$ as a function of the number of training epochs. Each metric is shown in three replicates corresponding to three randomly initiated training procedures.

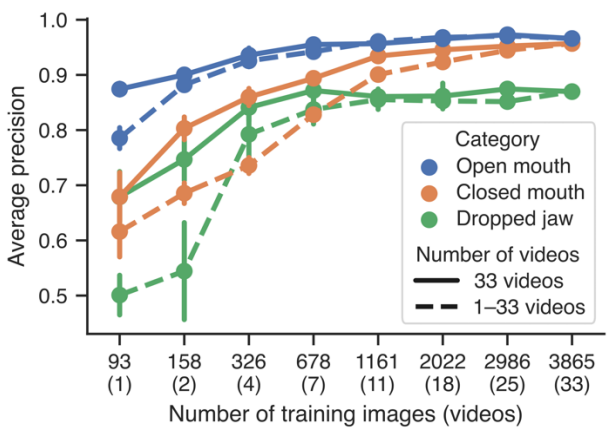

**Fig. 5.** Points plot showing the average precision on the validation dataset as a function of the number of training images. The dashed lines show models trained using datasets for which the number of training videos was subsampled. The error bars show the standard deviation of the three model replicates.

To test if the model's accuracy saturated with respect to the number of training images, we used the validation set to estimate the accuracy of models trained on subsets of training images (**Fig. 5**). Increasing the training set size from 93 images to 678 saw the sharpest increase in $AP_{50}$ values of the three categories: open mouths improved from 87.4% to 95.5%, closed mouths from 67.9% to 89.4%, and dropped jaws from 67.9% to 87.2%. Increasing the training set further to 3,865 saw only small or no improvements for the open mouth and dropped jaw categories. However, the closed mouth category $AP_{50}$ increased by 6.3 percentage points to 95.7%.

We repeated the subsampling experiments and progressively limited the number of videos from which training images were sourced (**Fig. 5**). At any training set size, this resulted in lower $AP_{50}$ values compared to sourcing images from all 33 videos, highlighting the need for having a large set of distinct videos in the training dataset. This trend was most pronounced for the closed mouth category, evidenced by differences of 6.6 and 3.3 percentage points when 678 and 1,161 images were sourced from 7 and 11, rather than from the full set of 33 training videos, respectively.

Subjecting the detection model to the test dataset, collected independently from the training and validation datasets, the performance for the open and closed mouths categories remained high with $AP_{50}$ values of 95.8% and 93.2%, respectively. Dropped jaws were detected with an $AP_{50}$ of only 70.0%, however the four test videos contained relatively few examples of dropped jaws, which may have contributed to the relatively imprecise estimate of the detection accuracy for this category.

### 4.2 Tracking by detection

Tracking accuracy of the fish head detection model in conjunction with a tracking-by-detection algorithm was evaluated using an independent test dataset of 100 manually annotated fish tracks. The mean tracking association accuracy was 97.7% (**Fig. 6**), meaning the tracking algorithm maintained the identity consistently over time without identity switches except in few cases where incorrect identity was assigned to the first part of the track (**Fig. 7**). The mean open and closed mouth recall values were 93.1% and 87.2%, while the mean precision values were 96.1% and 90.6%, respectively (**Fig. 6**). Most misclassification errors (92.0%) happened at the transition from an open to a closed mouth or from a closed to an open mouth (**Fig. 7**). These errors are to some extent unavoidable as the detection model tries to predict a categorical outcome to a continuous process of mouth closing and opening.

### 4.3 Estimating ventilation rate

We used our method to predict ventilation rates of the fish in the 100 fish tracks test dataset and compared the results with the ground truth, which was calculated from the manual annotations of the fish mouth states. The mean ground truth ventilation rate ranged from 28 to 150 open-closed cycles per minute with a median value of 103. Four fish never closed their mouth and our model correctly identified all four. For the remaining 96 fish, the predicted ventilation rates agreed well with the ground truth data yielding a Pearson correlation coefficient of 0.82 and a mean absolute error (MAE) of 7.15 (**Fig. 8a**). The proposed method cannot re-identify individual fish in a population and thus it is ideally suited for estimating population-level (i.e. fish pen) ventilation rates. When estimating the mean ventilation rate of each of the ten fish pens (rather than of individual fish), the mean absolute error improved to 3.58. The 96 fish spent more time with open mouths than with closed as evidenced by the median ground truth durations of open and closed mouth sequences of 11 and 6 frames (0.37 and 0.2 seconds), respectively. Our method predicted these durations with Pearson correlation coefficients of 0.84 and 0.90 and MAE of 2.04 and 0.94 frames, respectively (**Fig. 8b–c**).

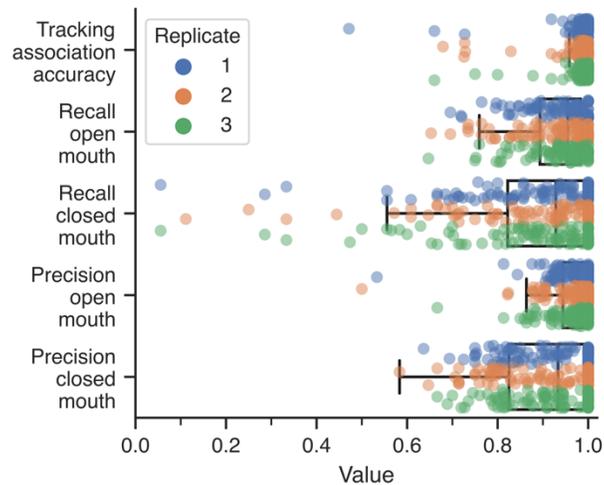

**Fig. 6.** Box plots showing the distribution of different tracking accuracy metrics across the 100 fish tracks. The whiskers were drawn to the farthest data points within 1.5 × interquartile range from the third quartile.

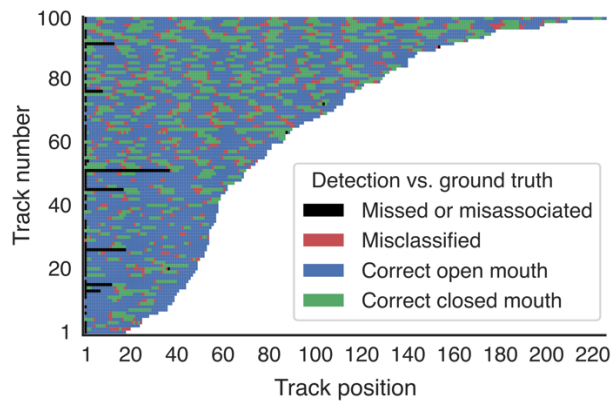

**Fig. 7.** Fish tracks coloured by the agreement between detections and ground truth. All red, blue, and green detections were correctly associated across frames without identity switches. For each track, detections from the model replicate with the lowest agreement with ground truth are shown.

As the final test, we used our method to provide a complete analysis of four videos from four different fish pens: two identified by salmon-industry professionals as containing fish with normal ventilation rates (Normal 1 and Normal 2) and two containing fish with elevated ventilation rates (High 1 and High 2). The number of fish detected per minute of footage varied from 925 to 201 fish (**Table 2**), due to differences in stocking densities of fish in pens and variations in water turbidity and camera angle. The proportion of fish for which at least one complete open-closed mouth cycle was observed ranged from 19% to 29% across the four videos. After quality control, an average of 64% of these fish were retained for ventilation rate estimation. Our method was able to detect the differences between the two pens with normal ventilation rate levels, estimated at 88.4 and 88.5 open-closed cycles per minute, and the two pens with elevated levels, estimated at 94.7 and 112.5 ($P < 0.01$ for all four comparisons, two-sided Mann–Whitney U test, **Table 2** and **Fig. 9**).

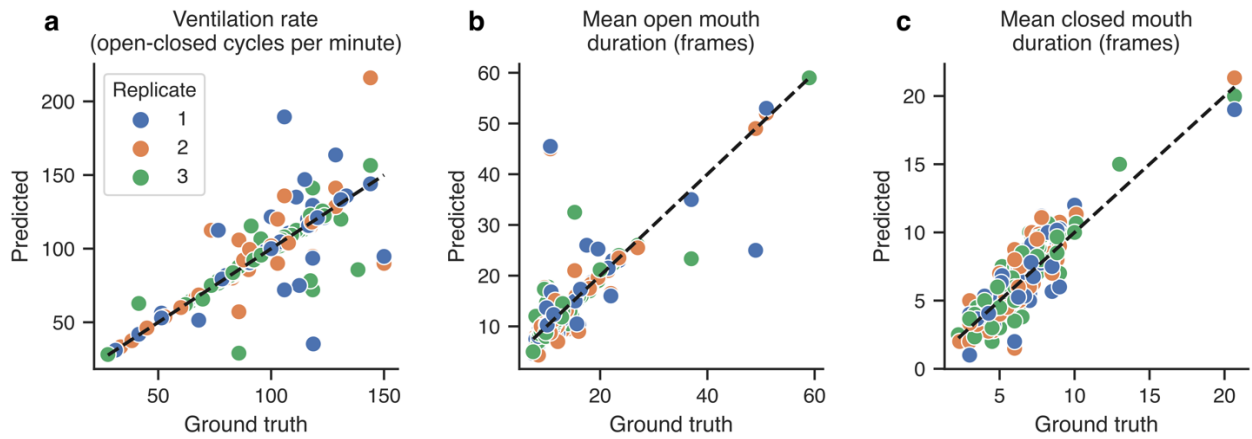

**Fig. 8.** Scatter plots showing the agreement of the predicted and ground truth ventilation rates (**a**), open mouth durations (**b**), and closed mouth durations (**c**). The dashed lines show perfect predictions.

**Table 2.** A complete analysis of four videos using the proposed method.

| Fish pen | Video length (mm:ss) | All fish | Fish with dropped jaws | Fish with ≥ 1 complete ventilation cycle | After quality control[a] | Median ventilation rate ± s.d.[b] | Mann–Whitney U test P value | |
|---|---|---|---|---|---|---|---|---|
| | | | | | | | Compared to High 1 | Compared to High 2 |
| **Normal 1** | 15:02 | 3,018 | 38 | 888 | 481 | 88.4 ± 0.6 | $9.2 \times 10^{-3}$ | $1.9 \times 10^{-10}$ |
| **Normal 2** | 15:06 | 7,535 | 85 | 1,519 | 891 | 88.5 ± 1.3 | $6.2 \times 10^{-6}$ | $4.1 \times 10^{-23}$ |
| **High 1** | 04:48 | 1,970 | 2 | 470 | 307 | 94.7 ± 0.0 | | |
| **High 2** | 01:03 | 971 | 1 | 180 | 138 | 112.5 ± 0.0 | | |

[a] Quality control consisted of discarding fish that contained […]-open-*closed*-open-[…] sequences.
[b] Standard deviation (s.d.) of the three model replicates.

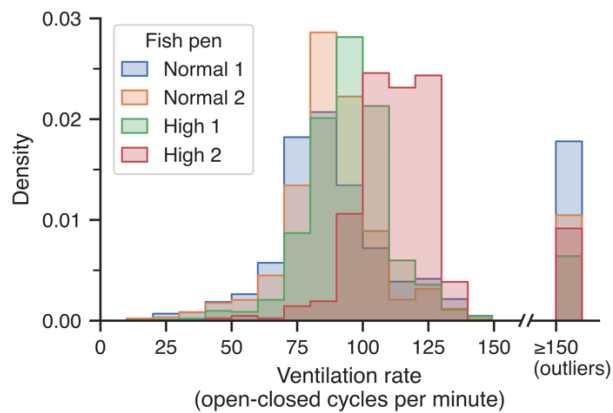

**Fig. 9.** Histograms showing the distributions of predicted ventilation rates in the four test videos.

### *4.4 Robustness to detection and tracking failures*

Robustness of the proposed method to missed detections and tracking association errors was evaluated using the four videos of which two were recorded in fish pens containing fish with normal ventilation rates and two in fish pens with elevated ventilation rates. We artificially corrupted detected fish tracks (in five replicates) and compared the estimated median ventilation rates with the uncorrupted estimates for each fish pen. The proposed method performed robustly across different corruption incidence levels (**Fig. 10**). When all detected tracks were corrupted with one to three missed detections, the difference in the median ventilation rate (ΔmVR) ranged from 0.0–0.7 open-closed cycles per minute across the four fish pens, as most of the introduced missed detections were correctly imputed (**Fig. 10a**). Increasing the corruption severity to two adjacent missed detections, which can no longer be imputed, resulted in shorter and fewer tracks, but saw only a small increase in the variability of the estimates with ΔmVR of 0.0–1.2 when 50% of the detected tracks were corrupted and ΔmVR of 0.0–1.5 after corrupting all tracks (**Fig. 10b**). Corrupting detected tracks with one to three tracking association errors (identity switches) resulted in shorter tracks and inclusion of multiple ventilation rate estimates for the same individual. The variability of the estimates remained low with ΔmVR of 0.0–1.0 when 50% of the detected tracks were corrupted and ΔmVR of 0.0–1.9 after corrupting all tracks (**Fig. 10c**). Finally, the statistically significant differences between the two fish pens with normal and the two fish pens with high ventilation rates could be detected even after introducing missed detections, adjacent missed detections, and tracking identity switches in 99%, 91%, and 100% of all the simulated scenarios, respectively, highlighting robustness of the proposed approach.

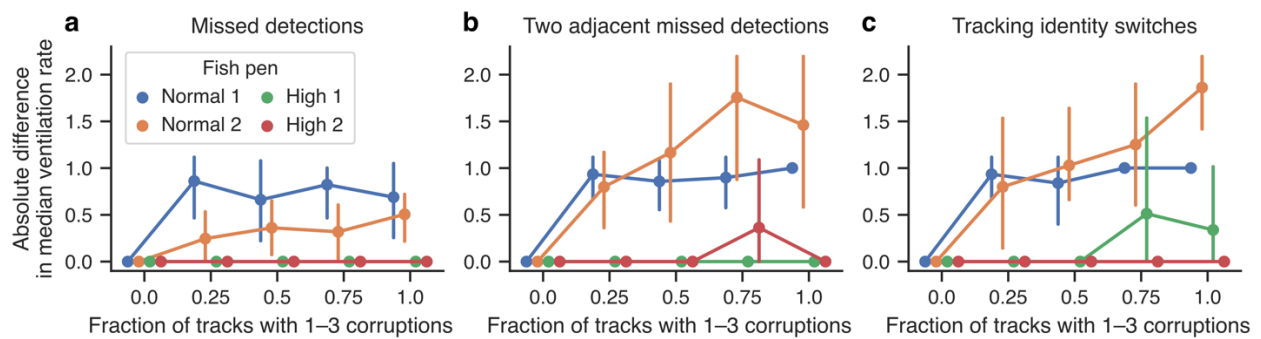

**Fig. 10.** Points plots showing robustness to missed detection (**a**), adjacent missed detections (**b**), and tracking identity switches (**c**). The error bars show the 95% confidence interval of five simulated corruptions.

*4.5 Processing speed and scalability*

The fish head detection and tracking model was capable of real-time processing with a speed of 31.7 ms per frame on average (NVIDIA A100 GPU). Approximately half of this time (14.5 ms) was occupied by the inference with the detection model and by moving the data between the GPU and CPU, while the remaining 17.2 ms was the time dedicated to tracking. Notably, tracking speed could be improved by avoiding the camera motion compensation algorithm, which constituted 16.8 ms of the tracking time. The computational cost of calculating the ventilation rate from the detection and tracking results was 2.1 ms per frame on average (Apple M2 chip).

We experimented with smaller YOLOv8 architectures (nano, small, and medium architectures comprising 3.2, 11.2, and 25.9 million parameters, respectively) and found satisfactory performance across our benchmarks (**Fig. 11**) with reduced computational times of 10.6 ms, 10.7 ms, and 12.6 ms, respectively. In addition, we used NVIDIA's TensorRT library to run the inference using the large YOLOv8 model, which reduced the running time to 13.2 ms and 7.2 ms (down from 14.5 ms), depending on whether 32-bit or 16-bit floating point precision was used. Overall, these results show that there are many possibilities for optimising the computational cost of our method depending on the specific needs of the aquaculture industry.

The most significant computational savings could be achieved by reducing the frame rate. However, due to the short period of time Atlantic salmon swim with their mouth fully closed (median value of 6 frames in our 100 fish tracks dataset), reducing frame rate comes at a considerable loss of accuracy. When we downsampled the 100 fish tracks dataset to 15 frames per second, the Pearson correlation coefficient between the predicted and ground truth ventilation rates decreased to 0.58 (down from 0.82 when 30 frames per second were used). Nonetheless, even at the lower frame rate the differences between the two videos with normal and the two videos with high ventilation rates were still detectable ($P < 2 \times 10^{-5}$ for all four comparisons, two-sided Mann–Whitney U test), highlighting that the model could be of practical utility even at 15 frames per second.

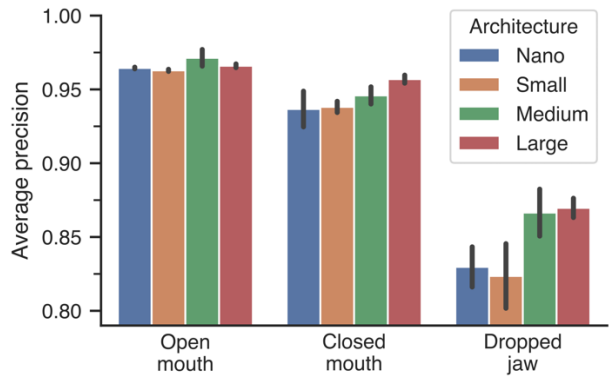

**Fig. 11.** Bar plot comparing the accuracy of different YOLOv8 model architectures. The error bars show the standard deviation of the three model replicates.

# 5 Discussion

We have developed a computer vision method to estimate ventilation rates from videos recorded in commercial sea fish farms, providing the aquaculture industry with a highly relevant, non-intrusive, and quantitative tool for monitoring ventilation rates, a vital measure of fish health and welfare (An et al., 2021). The overall accuracy of our approach stems from accurate detections of fish heads and accurate classification of the mouth states (open or closed). Therefore, we placed significant emphasis on the annotation of the training data. First, we minimised the occurrence of false positive training examples by removing all fish with narrow mouth openings. Second, to eliminate false negative examples, we blurred heads of fish located in the background of training images. Third, we collected training images from videos recorded under diverse environmental conditions (water turbidity, fish size, stocking density) and technical variables (colour cast, camera angle, image quality). Consequently, the fish head detection model correctly located fish with open and closed mouths with a high degree of accuracy. Coupled with robust tracking performance, our model consistently provided reliable estimates of ventilation rates.

The method was developed and evaluated using videos captured with cameras that are already integrated into the existing infrastructure at many fish farms (Raje, 2024), therefore, we do not foresee that the technical specifications of the video cameras as a significant barrier to industry adoption. Utilising existing infrastructure minimises the need for additional investment, which is limited to the computational resources to process the recorded videos. An important consideration for integrating the method into fish farm operations is that the video footage for the fish head detection model must be recorded with the camera oriented sideways, which can interfere with other tasks such as feeding or net cleaning. Nevertheless, we anticipate that even quantifying ventilation rates once a day for each pen would provide significant value to fish farm management.

Our method is highly scalable and designed to analyse large quantities of fish to provide population-level estimates of ventilation frequencies. This in turn ensures the method's robustness to detection and tracking failures. Missed detections and identity switches do not cause incorrect estimates as they merely decrease the number, and the length of fish tracks available for estimating the ventilation rates. While misclassified mouth states can affect the ventilation estimates, our results demonstrate that in practical applications, the high classification performance of the fish head detection model minimises their adverse impact and ensures reliable estimates of ventilation rates. Based on our data, monitoring the complete open-closed mouth cycle of just 500 fish provides a robust estimate of ventilation rates within a single pen. We found that analysing ventilation rates for a larger number of fish yielded minimal improvement in estimating the pen's median value. Although typical stocking densities of Atlantic salmon are at a ratio of 1.5% fish to 98.5% water (15 kg/m³) or less (Huon Aquaculture, 2024; Mowi, 2024; Tassal, 2022), salmon typically aggregate in areas of the pen with the most favourable water conditions (Compassion in World Farming, 2019). As a result, we estimate that 15 minutes of video recording is sufficient, provided that the camera position is optimised—a process that can be automated in many commercial aquaculture operations. Therefore, our method can analyse up to 120 fish pens within 24 hours using a single GPU and can be further scaled by processing multiple videos in parallel, depending on the GPU's memory capacity or by utilising multiple GPUs. The overall cost of implementing our method is comprised of a GPU-accelerated workstation and its electricity consumption, making it a practical and cost-effective solution even for large-scale aquaculture operations.

A potential bottleneck of the proposed method is that ventilation rates are estimated using only two mouth states—open or closed. While this approach performed well for videos recorded at 30 frames per second, lower frame rates would reduce its ability to detect closed mouth states due to the short duration the fish typically keep their mouth closed. Additionally, the high frame rate increases bandwidth requirements and electricity consumption. While we demonstrated robustness to missed detections, identity switches, and misclassified mouth states when calculating the median ventilation rate of a sufficiently large fish population, the method may be less suitable for small populations, such those kept in laboratory fish tanks. A model capable of continuous mouth state detection, such as keypoint detection, could improve performance at lower frame rates and enhance adaptability to smaller fish populations.

Fish ventilation rates can change rapidly, due to shifts in environmental conditions (Hvas & Oppedal, 2019) or stress from jellyfish (Pitt et al., 2024) and predator incursions into the pen (Quick et al., 2004), or slowly due to seasonal variations or the onset of diseases with slow progression such as those caused by pathogens affecting gill health (Oldham et al., 2016). Monitoring gill health is currently a laborious and time-consuming procedure that requires handling and sedating fish, which causes additional stress. At present, there is no efficient way to predict which pens should be prioritised for these manual health checks. Since gill disease impairs respiration (Bowden et al., 2022), our method offers a potential solution by improving the planning of manual health assessments. This could significantly enhance fish health and welfare while also reducing the costs associated with monitoring gill health.

As the demand for aquaculture production grows (FAO, 2022), and competition for aquatic space increases (Lee et al., 2021), the aquaculture industry's expansion will likely need to move further offshore. This shift may offer several advantages, including reduced environmental impacts due to greater water movement and improved fish health (Morro et al., 2022). However, operating at a greater distance from shore presents significant challenges, such as reduced accessibility due to weather conditions and increased costs associated with attending the farms to examine fish health in person. To address these challenges, comprehensive monitoring tools will be essential, enabling fish health personnel to plan manual inspections more efficiently and effectively. Computer vision and artificial intelligence approaches, like the method proposed here, will become indispensable tools facilitating the expansion of aquaculture into offshore environments.

## 6 Conclusions

We have developed a computer vision method to monitor ventilation rates in sea fish farms, offering significant advantages to the aquaculture industry. The method was specifically designed and evaluated using videos recorded in a production environment of commercial sea fish farms, demonstrating its suitability to practical applications. Our approach could reduce costs and animal handling associated with routine health checks through the identification of pens in which fish exhibit signs of increased respiratory effort, aiding in the prioritisation of pens for inspections. This early detection capability not only improves operational efficiency, but also has a positive impact on fish health and welfare. By enabling quantitative and long-term monitoring of ventilation rate—a critical physiological trait linked to various environmental and health conditions—our method has wide applicability and the potential to transform fish welfare monitoring in finfish aquaculture.

In future work, we plan to explore the transferability of our method to different fish species and experiment with lightweight neural architectures that would enable ventilation rate monitoring on edge devices (Restrepo-Arias et al., 2024). Implementing edge computing would significantly reduce bandwidth requirements, a critical consideration as the aquaculture industry expands further offshore.

## CRediT authorship contribution statement

Lukas Folkman: Conceptualization, Formal analysis and investigation, Data curation, Methodology, Software, Writing - original draft preparation, Writing - review and editing. Quynh Le Khanh Vo: Conceptualization, Data curation. Colin Johnston: Conceptualization, Data curation. Bela Stantic: Conceptualization, Funding acquisition, Writing - original draft preparation, Writing - review and editing. Kylie A. Pitt: Conceptualization, Funding acquisition, Writing - original draft preparation, Writing - review and editing.

## Declaration of competing interest

The authors declare that they have no known competing financial interests or personal relationships that could have appeared to influence the work reported in this paper.

## Acknowledgements

We thank Tassal for their support during the data collection. This research was funded by the Blue Economy Cooperative Research Centre, established and supported under the Australian Government's Cooperative Research Centres Program, grant number CRC20180101. We gratefully acknowledge the support of the Griffith University eResearch Service & Specialised Platforms Team and the use of the High Performance Computing Cluster "Gowonda" to complete this research.

## Data availability

Due to the sensitive nature of the datasets collected for this research, the datasets cannot be shared.